\documentclass[runningheads]{llncs}

\usepackage{eccv}

\usepackage{eccvabbrv}

\usepackage{graphicx}
\usepackage{booktabs}

\usepackage[accsupp]{axessibility}  %

\usepackage[pagebackref,breaklinks,colorlinks,citecolor=eccvblue]{hyperref}
\usepackage[dvipsnames]{xcolor}

\usepackage{caption, multirow, overpic, textpos}
\usepackage[font=footnotesize]{subcaption}
\usepackage{array}
\usepackage{amsmath}
\usepackage{colortbl}
\usepackage{booktabs} 

\usepackage[cjk]{kotex}

\usepackage{wrapfig}

\newcolumntype{x}[1]{>{\centering\arraybackslash}p{#1pt}}
\newcolumntype{y}[1]{>{\raggedright\arraybackslash}p{#1pt}}

\definecolor{deemph}{gray}{0.6}
\newcommand{\gc}[1]{\textcolor{deemph}{#1}}
\definecolor{baselinecolor}{gray}{.9}
\newcommand{\baseline}[1]{\cellcolor{baselinecolor}{#1}}

\newcolumntype{z}[1]{>{\raggedleft\arraybackslash}p{#1pt}}

\usepackage{xspace}
\newcommand{\oursfull}{Learning with Unmasked Tokens\xspace}
\newcommand{\ours}{LUT\xspace}

\definecolor{darkergreen}{RGB}{21, 152, 56}

\definecolor{red2}{RGB}{252, 54, 65}

\usepackage{orcidlink}

\begin{document}

\title{Learning with Unmasked Tokens Drives \\ Stronger Vision Learners} 

\titlerunning{Learning with Unmasked Tokens Drives Stronger Vision Learners}

\author{Taekyung Kim$^\star$ \and Sanghyuk Chun \and Byeongho Heo \and Dongyoon Han\thanks{Equal contribution}}

\authorrunning{Kim et al.}

\institute{NAVER AI Lab}

\maketitle

\begin{abstract}
Masked image modeling (MIM) has become a leading self-supervised learning strategy. MIMs such as Masked Autoencoder (MAE) learn strong representations by randomly masking input tokens for the encoder to process, with the decoder reconstructing the masked tokens to the input. However, MIM pre-trained encoders often exhibit a limited attention span, attributed to MIM's sole focus on regressing masked tokens only, which may impede the encoder's broader context learning. To tackle the limitation, we improve MIM by explicitly incorporating unmasked tokens into the training process. Specifically, our method enables the encoder to learn from broader context supervision, allowing unmasked tokens to experience broader contexts while the decoder reconstructs masked tokens. Thus, the encoded unmasked tokens are equipped with extensive contextual information, empowering masked tokens to leverage the enhanced unmasked tokens for MIM. As a result, our simple remedy trains more discriminative representations revealed by achieving 84.2\% top-1 accuracy with ViT-B on ImageNet-1K with 0.6\%p gain. We attribute the success to the enhanced pre-training method, as evidenced by the singular value spectrum and attention analyses. Finally, our models achieve significant performance gains at the downstream semantic segmentation and fine-grained visual classification tasks; and on diverse robust evaluation metrics. Code is available at {\small \url{https://github.com/naver-ai/lut}}.
\end{abstract}    
\section{Introduction}
\label{sec:intro}

Triggered by successful transitions of Transformer~\cite{vaswani2017attention} into vision domains \cite{dosovitskiy2020vit, carion2020detr}, a plethora of effective training strategies for Transformer have emerged \cite{touvron2021deit,chen2021mocov3, caron2021emerging,mae,touvron2022deit}. Recent advances in masked image modeling (MIM) \cite{Bao2021beit, zhou2021ibot, mae, xie2021simmim, dtm} noticeably show great success in self-supervised learning (SSL) of Vision Transformers (ViT) by transferring the knowledge of masked language modeling \cite{devlin2018bert}. Conceptually, MIM tasks consist of two parts: randomly masking out a part of inputs (\eg, 75\% of input pixels) and predicting the masked inputs by the decoder. This simple strategy enables a model to learn strong representations through the challenging task.

However, MIM strategies often encounter challenges, such as local dependency on attention to understand the entire context of an image. For example, liu~\etal~\cite{liu2022exploring} revealed that Masked Autoencoder (MAE)~\cite{mae}, a state-of-the-art MIM method, exhibits shorter average attention distances. Furthermore, we observe that attention map patterns by MAE substantiate extremely local behavior (See Fig.~\ref{fig:attn_maps}) indeed. In other words, the MAE-trained attention mechanism integrates less information across all image pixels and tends to focus on specific input regions. This is presumably attributed to MIM-pretraining, primarily dedicated to predicting low-level pixel details (\eg, color or texture) without a comprehensive understanding of less-regional information (\eg, the input structure or shape).

We aim to understand the chronic shortage in a limited range of dependencies and how it affects MIM. We illustrate that vanilla MIM methods appear to lack broader-range dependency. Drawing from this, we identify a deficiency inside the vanilla MIM formulation and present a simple solution to its local dependency issue, demonstrating how it enhances MIM pre-training. Our proposed method \oursfull (\ours) for MIM complements the sub-optimal representation learning by offering broadly contextualized supervision to unmasked tokens through extracting general context from the entire pixels to strengthen unmasked tokens (for masked tokens to attend), which aims to learn more context-generalized representations for the encoder.

During training, \ours minimizes the discrepancy between the encoded general context representations and the sparse representation processed by the online learnable encoder from different views while performing MIM with a decoder. This ensures reinforcing more contextualized unmasked tokens for mask tokens to attend to. A general representation derived from all pixels can effectively utilize a highly augmented view, minimizing reliance on regional changes like color distortion to improve generalization. The learnable network encodes a sparse and unmasked view and matches it to the generalized representation, and the decoder reconstructs the masked pixels using the encoded features, similar to \cite{mae}. 
We presume that our strategy promotes the learning of the encoder by involving unmasked tokens in training, leveraging the target network's capability to encode a wide context for all pixels.

We verify the effectiveness of \ours by pre-training the ViT architectures~\cite{dosovitskiy2020vit} on the ImageNet-1K benchmark \cite{imagenet}. We do not solely evaluate our method by linear probing but by fine-tuning results. Given our method's weight on improving the baseline MIM, \ours-trained ViT-B/16 successfully improves linear evaluation (+2\%p) and fine-tuning (+0.6\%p) performance gains on ImageNet-1K over MAE. Our fine-tuning result also achieves comparable or outperformed ImageNet-1K validation accuracy (84.2\%) compared with other state-of-the-art methods. \ours can be transferred to the multiple fine-grained classifications and show distinguished transferability. \ours further shows superior transferability and tuning robustness on INaturalist datasets. We further transfer our pre-trained model to the semantic segmentation task on ADE20K~\cite{zhou2017ade20k} and show 48.6\% mIoU, a solid result. As another benefit, \ours successfully realizes robust training, which results in enhanced robustness results on two in-distribution benchmarks, five out-of-distribution benchmarks, and SI-Score \cite{si_score}.
\section{Preliminary}
\label{sec:motivation}
Despite MIM's strong performance, we claim it still lacks strong attention capability after pre-training, particularly for comprehensive region-wide dependency. The upcoming spatial attention map visualizations motivate our method. 

\subsection{MIM and Beyond}\label{sec3:mim}
We begin with a generalized formulation of MIM, addressing the limitation shown in the formulation.

\vspace{-0.3em}
\subsubsection{General formulation.}
Given an image from an augmented view $u$, we patchify the image into $N$ non-overlapping patches $U = \{\textbf{u}_i\}^N_{i=1}$. We randomly pick masked patches $\mathcal{M}$ with a high masking ratio $r \in (0,1)$, where $\mathcal{M} \subset \{1,2, ..., N\}$ and $|\mathcal{M}| = \lfloor r N \rfloor$.
We denote the masked image patches as $U_{m} = \{\textbf{u}_i:i\in\mathcal{M}\}$ and the unmasked patches as $U_{r} = \{\textbf{u}_i:i\in \{1,2, ..., N\}, i\notin\mathcal{M}\}$. 
The unmasked patches are fed into the encoder $f_\theta$ and become encoded tokens $T_e = f_\theta(U_{r})$. The encoded tokens are concatenated with mask tokens $\textbf{m}_i$ corresponding to the positions of $i$-th masked patches (entire patches $U_{m} \cup U_{r}$ can be fed into the encoder~\cite{xie2021simmim} or $U_{m}$ only is utilized~\cite{mae}). The only mask tokens predict the image patches through the decoder $d_\phi$. 
We denote i-th decoded mask token and input mask token as $\textbf{m}_i^d$ and $\textbf{m}_i$, where 
${T_d \cup \{\textbf{m}^d_i\}_{i\in\mathcal{M}}} =d_\phi(T_e\cup \{\textbf{m}_i\}_{i\in\mathcal{M}})$, informally. Here, $T_d$ is a set of decoded unmasked tokens. Now, the MIM pre-training objective is defined as:
\begin{equation}
\label{eq:mim_loss}
    \mathcal{L}_{\text{MIM}} = \sum_{i\in\mathcal{M}} || \textbf{m}_i^d - \textbf{u}_i||^2_2,
\end{equation}
where $m_i$ are shared for all the positions.

\subsubsection{MIM formulation itself falls short in learning broader contexts.} 
We argue that the MIM loss in Eq.~\eqref{eq:mim_loss} may not fully exploit the pre-training capability while it is effectively designed with simplicity. Specifically, owing to the scarcity of unmasked tokens $U_r$ in the encoder $f_\theta$, the formulation does not leverage complete image information; they are encoded through self-attention, being only attended by a few visual tokens. Therefore, the encoding may struggle with low contextualization due to sparse visual tokens from the query. 

\begin{figure*}[t]
     \centering
     \begin{subfigure}[b]{0.1125\linewidth}
         \centering
         \includegraphics[width=1\textwidth]{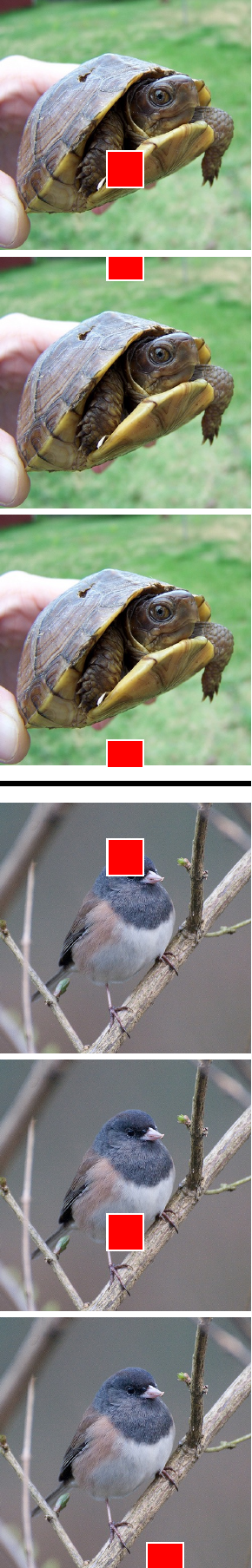}
         \vspace{-1.3em}
         \caption{Input}
         \label{fig:input}
     \end{subfigure}
     \begin{subfigure}[b]{0.348\linewidth}
         \centering
         \includegraphics[width=1\textwidth]{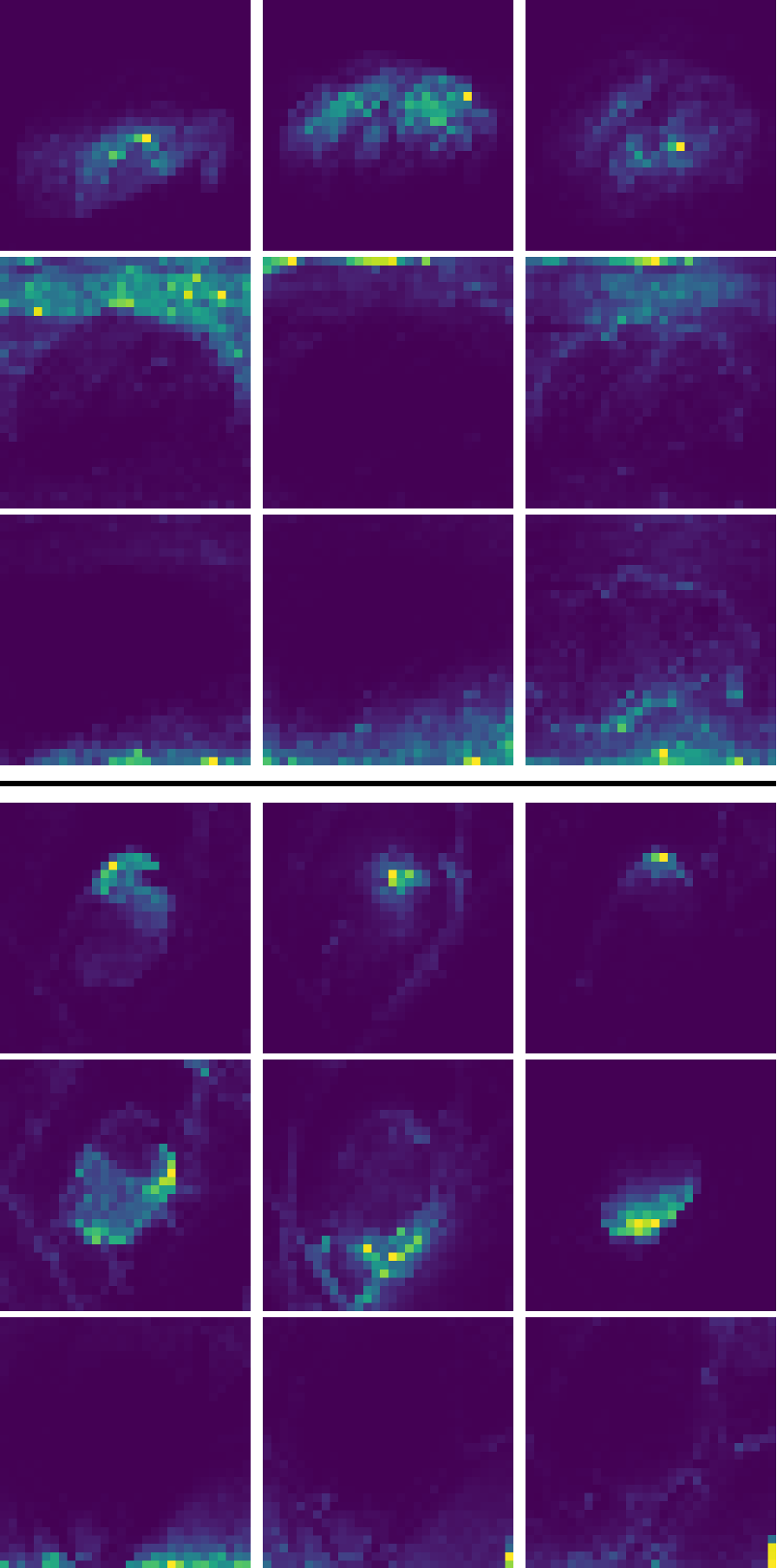}
         \vspace{-1.3em}
         \caption{MAE}
         \label{fig:attn_mae}
     \end{subfigure}
     \begin{subfigure}[b]{0.348\linewidth}
         \centering
         \includegraphics[width=1\textwidth]{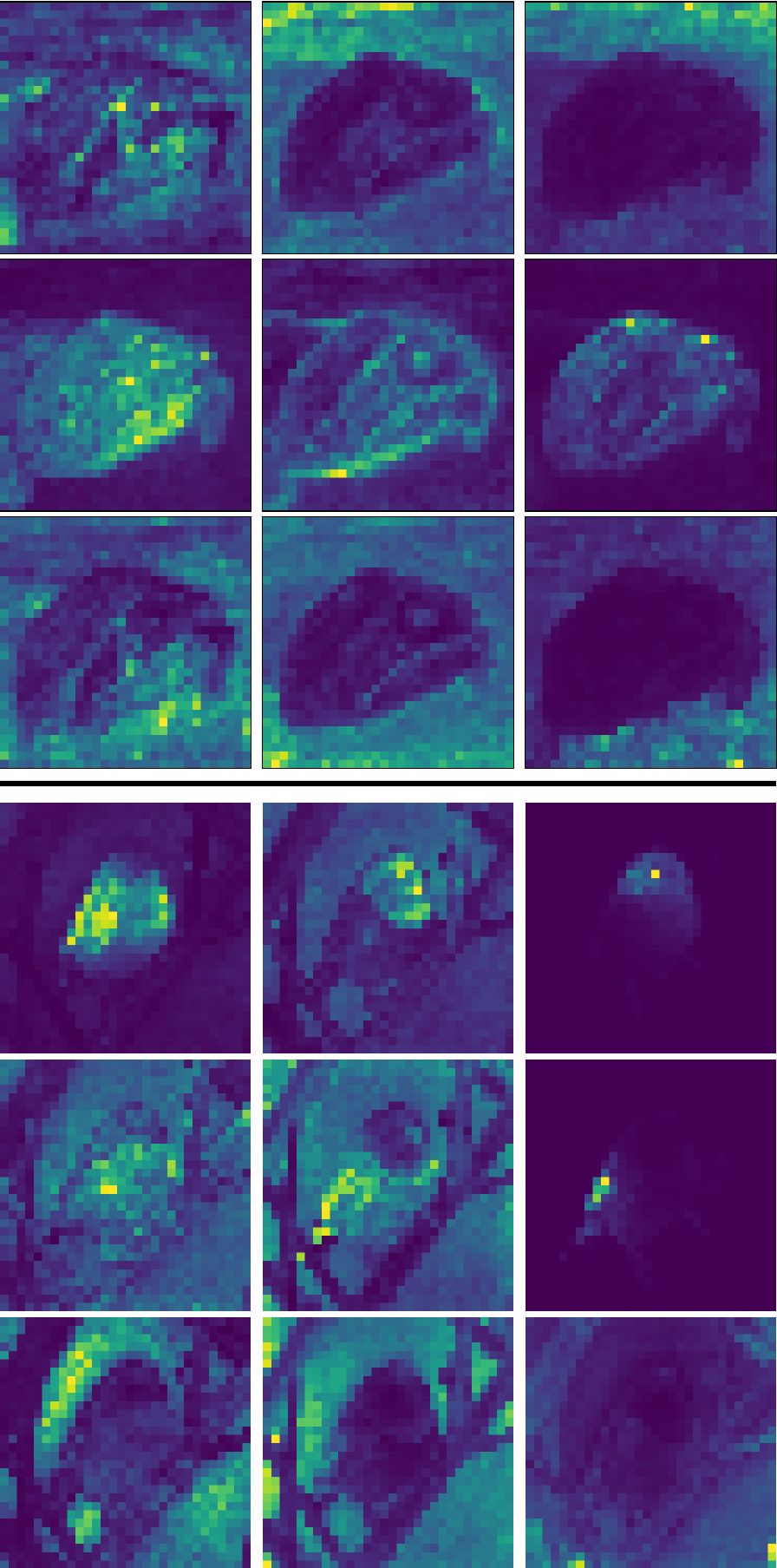}
         \vspace{-1.3em}
         \caption{\ours (Ours)}
         \label{fig:attn_ours}
     \end{subfigure}
    \vspace{-.5em}
        \caption{\textbf{Motivation - MAE may lack comprehensive region-wide attention.}
        We observed how attention appears differently in MAE  corresponding to given queries. (a) The first column denotes the example images with  (\textcolor{red}{\textbf{different queries}}) randomly picked patch indices. (b), (c) Every set of three columns represents the maps that are the most attended by different heads. The \textcolor{ForestGreen}{\textbf{turtle}} images have a foreground, and two upper and lower background queries; the  \textcolor{BlueViolet}{\textbf{bird}} images have two foreground queries (upper two rows) and one background query. MAE shows localized attention maps but fails to provide comprehensive coverage of either foreground or background.}
        \label{fig:attn_maps}
        \vspace{-1em} 
\end{figure*}

In the decoder $d_\phi$, we argue the restricted local options $T_e$ are only available for the mask tokens to attend to. The loss is primitively to provide limited supervision for masked tokens to reconstruct image patches from the constrained encoded information $T_e$ using the decoder. The mask tokens may eventually attend near visual tokens to reduce the loss, as we will observe in Fig.~\ref{fig:attn_mae}. 

\subsection{Motivation - attention map visualizations} 
Attention map visualizations qualitatively reveal how a model reacts to queries and reflects the deficiency of representations. We believe the range of attention may suggest the diversity of token dependency that closely links to the encoder's capability. Fig. \ref{fig:attn_mae} shows the attention maps concerning the given query in Fig. \ref{fig:input} by MAE \cite{mae}. We exploit self-attention in the last block for visualization in the official ViT-B/16 MAE and visualize maps with $480\times480$ images from ImageNet-1K. We observe MAE shows narrower highlighted regions for the given queries. Specifically, when a query is selected in the foreground (the 1st, 4th, and 5th rows), MAE only highlights the near patches of the given query; when a query is selected even in the background (the 2nd, 3rd, and 6th rows), we observe the same phenomenon, namely, MAE only focuses on the near patches of the given query. Based on this, we argue that MAE's attention lacks a broader range of dependency. This may incur a lack of a broad understanding of the entire foreground or background.
\begin{figure*}[t]
    \centering
    \includegraphics[width=.9\textwidth]{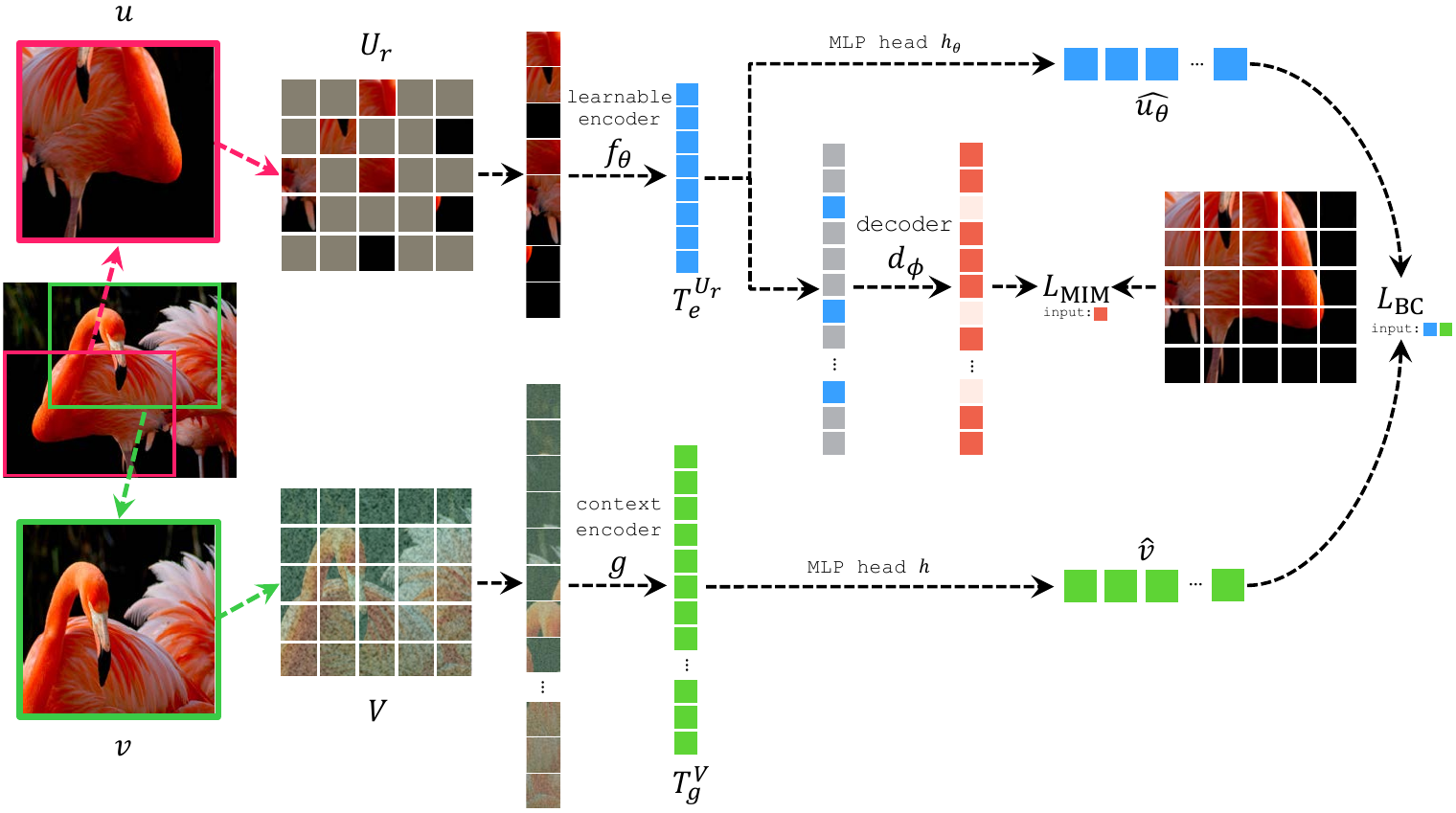}
    \vspace{-1em}
    \caption{
    \textbf{Framework overview.} Our method performs a masked image modeling with masked tokens, complemented by a context encoder that directs the learnable side, giving an augmented complete view for sparse unmasked tokens. We employ distinct, simple MLP heads on top of each encoder to match representations and avoid optimization collapse. %
    Under our macro concept, we opt for a simple choice where the additional context momentum encoder mirrors the online encoder, while we may alternatively use various options.
    We borrow a flamingo image from \texttt{n02007558} class in ImageNet-1K.
    }
    \label{fig:sec3_fra mework}
    \vspace{-1em}    
\end{figure*}

\section{Method}
\label{sec:sec3}
In this section, we introduce our proposed method - \oursfull (\ours) for Masked Autoencoder that addresses the local dependency issue. We opt for MAE~\cite{mae} as the baseline for our formulation.

\vspace{-1em}
\subsubsection{Our simple solution.}
Based on the arguments presented, we contend that Eq.~\eqref{eq:mim_loss} computes loss exclusively using reconstructed masked tokens while leaving unmasked tokens being trained implicitly, which may impede learning broad contexts effectively. As a remedy, we employ another loss expected to aid $\mathcal{L}_{\text{MIM}}$ by giving expansive supervision to unmasked tokens from the entire tokens:
\begin{equation}
\label{eq:mim_extended_loss}
    \mathcal{L}_{\text{ours}} = \sum_{i\in\mathcal{M}} || \textbf{m}_i^d - \textbf{u}_i||^2_2 + \alpha \mathcal{D}(T_e, g(U_{m} \cup U_{r})),
\end{equation}
where $\mathcal{D}(\cdot, \cdot)$ and $g(\cdot)$ denote a distance function and a context encoder. We straightforwardly give encoded comprehensive supervision from entire tokens to unmasked tokens $T_e$. During training, the expansively supervised $T_e$ contains extended token information so that mask tokens can leverage it. This potentially gives additional localization capability beyond what the baseline possesses. The options for choosing $\mathcal{D}$ and $g$ are indeed diverse, but we take the simplest way in the next section. Eq. \eqref{eq:mim_extended_loss} can involve both unmasked and mask tokens, but we focus on unmasked tokens to prevent learning collapse in mask tokens.

\vspace{-1em}
\subsubsection{Contextualized supervision.}
The crux of our solution lies in \textit{learning unmasked tokens actively} from a more comprehensive contextualization of entire visual tokens. We opt for the elements in the newly involved loss (dubbed broader contextualization loss $\mathcal{L}_{\text{BC}}$) in Eq.~\eqref{eq:mim_extended_loss}. First, for the context encoder $g$, we implement this by simply reusing the encoder $f_{\theta}$ to give the supervision back to $f_{\theta}$. It performs like a token-level regression between the encoders. 
We employ an efficient yet strong option, momentum networks~\cite{he2019moco, chen2021mocov3,grill2020byol,caron2020unsupervised}.
The architecture consists of a momentum encoder and MLP head, which is nearly identical to the learnable side. %

Additionally, we augment the entire image patches from $U$ to $V$ to enhance the generalization of the encoder and avoid collapse.
For the context latent features, the view $\textbf{v}$ is patchified into $V = \{\textbf{v}_i\}^N_{i=1}$, respectively.
Unlike the general MIM process, all the patches $V$ are encoded by %
$g$; we denote the encoded tokens as $T^V_g = g(V)$, where $T^V_g = \{t^V_i\}^N_{i=1}$.
Finally, the MLP head $h$ yields context representations $\hat{\textbf{v}}=h(\tilde{\textbf{v}})$, where $\tilde{\textbf{v}}$ can be $\tilde{\textbf{v}}=\frac{1}{N}\Sigma^N_{i=1} t^V_i$ (pooled) or each set of representations $\tilde{\textbf{v}}=T^V_g$.
Alternatively, using aligned tokens~\cite{cuturi2013sinkhorn} for $\hat{\textbf{v}}$ could benefit performance, but we simply use a pooled token.

We refer to the process as delivering \textit{contextualized supervision} to unmasked tokens, which encodes information from entire tokens to facilitate training through the broader supervision for unmasked tokens. Note that this supervision branch adds a single forward pass, but ours exhibits a slight
time increase of approximately 18\% - 0.46s over 0.39s (baseline) at
 each iteration.%

\vspace{-1em}
\subsubsection{Sparse unmasked tokens that learn broad contexts.}
Our encoding process
obtains regional representations $T^{U_r}_e = f_\theta(U_r)$ from sparsified tokens $U_r$.
Similar to computing general context representations, we aggregate the latent embeddings $\textbf{u}_\theta=\frac{1}{|T^{U_r}_e|}\Sigma_{t\in T^{U_r}_e} t$ through averaging. We follow the previous studies preventing training collapse by applying an MLP head $h_\theta$ to obtain $\hat{\textbf{u}_\theta}=h_\theta( \textbf{u}_\theta )$, forming architectural asymmetry to avoid collapse~\cite{grill2020byol,chen2021simsiam}.
\ours can be interpreted as utilizing masked tokens for MIM interacting with sparse visual tokens employed to condense expanded context information.

\vspace{-1em}
\subsubsection{On contextual discrepancies across views.}
We aim to provide broader contextualized supervision to unmasked tokens that correspond to the original view of the masked tokens. However, MIMs generally use random resized crop (RRC)~\cite{inceptionv1} for giving geometric variation; we argue that using RRC may not align with our intention and could hinder learning due to divergent views often providing narrower and limited shared information \cite{chen2021simsiam, touvron2022deit}.
Thus, we adopt simple resized crop (SRC) \cite{touvron2022deit} instead of RRC. We conjecture the latent features from unmasked tokens can be more reliably guided by the semantics from a broader context.
We will show that SRC harms MAE but improves \ours in \S\ref{sec:analysis}.

\vspace{-1em}
\subsubsection{Objective function.}
We finalize our objective by choosing the distance function $\mathcal{D}$ in Eq.~\eqref{eq:mim_extended_loss}. 
We apply the normalized $\ell_2$-distance for the feature distance (\ie, Cosine distance).
We have the aggregated context representation $\hat{v}$ and sparse one $\hat{\textbf{u}}_{\theta}$, and their $\ell_2$-normalized version $\bar{\textbf{u}}_{\theta}$ and $\bar{\textbf{v}}$, respectively.
Our broader contextualization loss $\mathcal{L}_{\text{BC}}$ computes the feature distance between normalized representations $\bar{\textbf{u}}_{\theta}$ and $\bar{\textbf{v}}$, formulated as
    $\mathcal{L}_{\text{BC}} = ||\bar{\textbf{u}_\theta} - \bar{\textbf{v}} ||^2_2.$
\ours is agnostic to the choice of distance function since the fundamental principle of it works regardless of the distance functions - the InfoNCE or Smoothed $\ell_1$ losses also show compatibility with \ours.
The final objective function is:
\begin{equation}
\label{eq:ours}
    \min_\theta\, \mathcal{L}_{\text{MIM}} + \alpha \mathcal{L}_{\text{BC}},
\end{equation}
where $\alpha$ controls the balance of the broader contextualization loss and the masked image modeling loss. The study on $\alpha$ gives the best fine-tuning performance with 0.25; however, it is insensitive to the choice of $\alpha$; for example, $\alpha$ = 0.25 and 0.5 shows only 0.1\% difference of fine-tuning performance using the ViT-B/16 backbone. We support all our design choices in the ablation studies in Table.~\ref{tab:ablation}. Our method is also applicable to SimMIM~\cite{xie2021simmim}-like methods with performance improvements (see Appendix for details).

\section{Related Work}
\label{sec:prior_arts}
Prior to transitioning to our experiments, we outline the distinctions between our work and closely related works. Several studies have been conducted recently employing multiple encoders, such as our online and target encoders. For example, a line of research excludes using additional data and employs an additional tokenizer module for reconstruction supervision. Drawing inspiration from the success of masked language modeling~\cite{devlin2018bert}, milestone MIMs~\cite{xie2021simmim,Bao2021beit,mae,zhou2021ibot,data2vec} have also gained attention for their ability to train discriminative representations by masking input tokens to reconstruct them during pre-training. 

First, iBOT~\cite{zhou2021ibot} jointly trains the online encoder and the online tokenizer. The main motivation is to align the full representations of multi-view instances among 12 different views while additionally performing masked feature reconstruction. Thus, iBOT needs multi-crops varying in diverse scales and augmentations. iBOT involves only mask tokens to learn target information, which may incur learning partial information, but we nevertheless speculate that leveraging multi-crops diminishes this issue. In contrast, our aim is to employ unmasked tokens to reinforce the understanding of longer-range context by offering complete information from a single view.

data2vec\cite{data2vec} performs patch-wise feature prediction via masked tokens. Except for the tokenizer, the framework is largely based on the BeiT framework~\cite{Bao2021beit}. Despite the target momentum features being generated from an entire image, we argue that data2vec still lacks long-range supervision to learn. Specifically, we presume only masked tokens contribute to MIM, suggesting attention between masked and unmasked tokens alone is insufficient for learning contextualized information. Therefore, the patch-wise regression to the token representations may not adequately establish strong neighboring dependencies. We conjecture this was eventually evidenced by inferior localization performance.
MSN\cite{msn} is a denoising-based method employing masks for denoise without reconstruction loss. Its focus on low-shot learning and prototype-based matching before supervision are major distinctions to our method. CMAE~\cite{cmae} and ConMIM~\cite{conmim} both focus on contrastive learning along with MIM. 
However, a major difference of ours stems from the primitive goal. The methods directly combine contrastive learning with all the visual patches recovered by particular reconstruction decoders. On the other hand, our method does not perform contrastive learning but targets to forge the unmasked tokens excluded from MIM training by directing them with comprehensive contextualized information.

\section{Experiment}
\label{sec:exp}
This section demonstrates our method by pre-training on ImageNet-1K, compared with competing SSL methods. We further fine-tune our models to confirm transferability. 
\subsection{ImageNet-1K Classification}
\subsubsection{Architecture.}
We use the standard Vision Transformer (ViT)~\cite{dosovitskiy2020vit} with a patch size of 16 for all experiments (\ie, ViT-B/16) to fairly compare with prior arts. We use the 8-layer transformer decoder following the MAE~\cite{mae}'s setup for masked image modeling. We further adopt online and context MLP heads on the top of the encoders to aggregate general context from representations; each consists of fully-connected layers with the embedding dimension of 4096, batch normalization layers \cite{ioffe2015bn}, and ReLUs \cite{krizhevsky2017alexnet} following the previous methods~\cite{grill2020byol,chen2021mocov3, caron2021emerging}. Note that \ours works even with symmetric heads. All the decoder and MLP heads are only used during training.

\vspace{-1em}
\subsubsection{Pre-training setup.}
We follow the identical ImageNet-1K~\cite{imagenet} pre-training protocol\footnote{We use the public codebase in \url{https://github.com/facebookresearch/mae}}~\cite{mae}. Our model is pre-trained for 1600 epochs with 40 warmup epochs, batch size of 4096, and input resolution of 224$\times$224. We use AdamW~\cite{loshchilov2018decoupled} with momentum (0.9, 0.999). The learning rate is set to $1.5\times10^{-4}$ with cosine learning rate decay~\cite{loshchilov2017sgdr}. We adopt a layer-wise learning rate decay of 0.65.
We set a mask ratio for Eq.~\eqref{eq:mim_loss} to 0.75, a momentum decay rate $\tau$ in the target network to 0.996, and the weight of our broader contextualization loss ($\alpha$ in Eq.~\eqref{eq:ours}) to 1.0 and 0.25 for the ViT-S/16 and ViT-B/16 architectures, respectively. We employ the simple resized crop~\cite{touvron2022deit} for geometric augmentation, color jittering, and the three augment~\cite{touvron2021deit} consists of Gaussian blur, grayscale, and solarization. All models are pre-trained using 8 V100-32GB GPUs.

\begin{table*}[t!]
\small
\tabcolsep=.6em
\centering
\caption{
\textbf{Comparisons with previous models on ImageNet-1K}. We compare \ours with the previous results that used vanilla Vision Transformer architectures. All models were pre-trained and fine-tuned on ImageNet-1K.%
We use the ViT-S/16, ViT-B/16, and ViT-L/16 architectures and a resolution of 224$\times$224. $^\dagger$ denotes the models pre-trained using multi-crop augmentation. $^\ddagger$ denotes our reproduction results. We highlight the best numbers (in boldface) and the second-best numbers (in underlined).
For a fair comparison, we do not compare methods using modules trained on extra data, such as CLIP \cite{clip} or VQGAN \cite{esser2021vqgan}. 
}
\label{tab:imagenet1k_sota}
\resizebox{1\linewidth}{!}{
\begin{tabular}{lccllll}
\toprule
\multirow{1}{*}{Method} 
& \multicolumn{1}{c}{} 
& \multicolumn{1}{c}{Supervision} 
&  ViT-S
&  ViT-B
&  ViT-L
&  ADE20K \\
\midrule
\textit{\gc{Sup models}}\\
DeiT~\cite{touvron2021deit} & ICML 2021  & Label  & 79.9 & 81.8 & \;\; - & \;\; - \\ %
DeiT-III~\cite{touvron2022deit} & ECCV 2022  & Label & 81.4 & 83.8 & 84.2 & 49.3\\
Cosub~\cite{touvron2023co} & CVPR 2023  & Label & 81.5 & \textbf{84.2} & 85.3 & 49.3\\
\midrule
\textit{\gc{Self-sup models}}\\
MoCo v3 \cite{chen2021mocov3} & ICCV 2021  &  Pixel   & 81.4 & 83.2 & 84.1 & 47.3\\
DINO \cite{caron2021emerging} & ICCV 2021  &  Pixel & 81.5$^\dagger$ & 82.8$^\dagger$ & \;\; - & 46.8\\
SplistMask~\cite{splitmask} & arXiv 2021 & Pixel+Feature &  81.5 & 83.6 & \;\; - & 46.8\\
iBOT \cite{zhou2021ibot} & ICLR 2022   &  Feature & \textbf{82.0} & 84.0$^\dagger$ & 84.8$^\dagger$ & \textbf{50.0}$^\dagger$\\
MAE \cite{mae} & CVPR 2022   & Pixel &  81.4$^\ddagger$  & 83.7$^\ddagger$ & 85.6$^\ddagger$ & 48.1\\
SimMIM \cite{xie2021simmim} & CVPR 2022  &  Pixel  &\underline{81.9$^\ddagger$} & 83.8 & \;\; - & \;\; - \\
MaskFeat \cite{wei2022masked} & CVPR 2022   & Feature & \;\; - &  84.0 & 85.7 & \;\; - \\
ExtreMa \cite{extrema} & TMLR 2022   & Feature & 81.8 & 83.7 & \;\; - & 47.9 \\
data2vec \cite{data2vec}  & ICML 2022  & Feature & 81.8$^\ddagger$ & \underline{84.1}$^\ddagger$ & \textbf{86.6} & 48.3$^\ddagger$ \\
SemMAE \cite{li2022semmae} & NeurIPS 2022  &  Pixel  & \;\; -  & 83.3 & \;\; -& 46.3 \\
SdAE~\cite{sdae} & ECCV 2022  & Pixel & \;\; - & 84.1$^\dagger$ & \;\; - & 48.6$^\dagger$ \\
MSN \cite{msn} & ECCV 2022   & Feature & \;\; -  &  83.4 & \;\; - & \;\; - \\
BootMAE~\cite{bootmae}  & ECCV 2022    & Pixel+Feature & \;\; - & \textbf{84.2} & 85.9 & 49.1 \\
CAN \cite{can} & arXiv 2022 & Pixel & \;\; - & 83.6 & 84.7 & \;\; - \\
ConMIM~\cite{conmim} & ICLR 2023   & Dictionary & \textbf{82.0} & 83.7 & 85.5 & 46.0 \\
SIM \cite{sim} & CVPR 2023 & Feature & \;\; - & 83.8 & \;\; - & \;\; - \\
HPM~\cite{hpm} & CVPR 2023 & Pixel & \;\; - & \textbf{84.2} & 85.8 & 48.5 \\  
MIRL~\cite{mirl} & NeurIPS 2023 & Pixel & \;\; - & 84.1 & 85.4 &  \\  
CrossMAE~\cite{crossmae} & arXiv 2024 & Pixel & 79.3 & 83.7 & 85.4 &  \\  
\midrule
Ours  & - &   Pixel & \bf{82.0}& \textbf{84.2} & \underline{86.0} & \underline{49.5}\\ 
\bottomrule
\end{tabular}
}
\vspace{-1.5em}
\end{table*}

\subsubsection{Results.}
We compare our method with previous SSL methods~\cite{chen2021mocov3, caron2021emerging, mae, xie2021simmim, zhou2021ibot, wei2022masked, extrema, data2vec, li2022semmae, sdae, msn, bootmae, can, conmim, sim, hpm}. Table~\ref{tab:imagenet1k_sota} shows the evaluation results on the ViT-S/B/L backbones.
Our \ours achieves an 82.0\%, 84.2\%, and 86.0\% top-1 accuracy on ViT-S/16, ViT-B/16, and ViT-L/16, which improves 0.6$\%$p, 0.6$\%$p, and 0.4$\%$p over the baseline, respectively. Moreover, \ours outperforms other self-supervised learning methods by a large margin except for some masked feature models. This comes to a head with a smaller ViT-S/16, where most of the results are saturated, but this is presumably due to the low capability of the backbone and the high flexibility of masked feature models. \ours would take advantage of further improvements using masked feature models as the baseline. The results highlight the efficacy of our proposed broader contextualized supervision in enhancing MIM, which showcases its significant potential for further improvements.

\vspace{-1em}
\subsubsection{Computational costs.}
Our method includes extra computation from forward inference with images, so there is a slight increase in computational demands, as mentioned above. However, our method achieves a top-1 accuracy of 83.6\% at 400 epochs, which matches MAE's accuracy at 1600 epochs, despite our significantly shorter GPU wall time. Specifically, our method takes 119 hours to complete 400 epochs of training, which is roughly half the training time of MAE's 1600 epochs, which requires 223 hours. \subsection{ADE20K Semantic Segmentation}
To validate the transferability to dense prediction tasks, we evaluate semantic segmentation performances on ADE20K~\cite{zhou2017ade20k}. 
We follow the standard training protocol~\cite{mae}; the models are fine-tuned for 160K iterations using UperNet~\cite{upernet} with a batch size of 16 and a resolution of 512$\times$512. Other detailed hyper-parameters for training are listed in Appendix.
The rightmost column in Table \ref{tab:imagenet1k_sota} shows the mIoU performance comparison. \ours also outperforms the competing methods, including SSL and supervised learning methods. This outcome can be attributed to the improved dense prediction capability.

\begin{table}[b]
\small
\caption{
\textbf{Transfer learning results on iNaturalists}.
We present the end-to-end fine-tuning accuracies on iNaturalist 2018, iNaturalist 2019, and mini iNaturalist 2021~\cite{inaturalist}. We report the best results along with the mean ± std of the set of accuracies obtained from grid searches for each method. $^\dagger$ denotes the models pre-trained using multi-crop augmentations. Our method consistently outperforms the competitors in terms of the best accuracies, further showcasing remarkable tuning robustness. 
}
\label{tab:transfer_cls_more}
\vspace{-1em}
\tabcolsep=0.7em
\centering
\resizebox{0.75\linewidth}{!}{
\begin{tabular}{l c c c}
\toprule
Method & iNat 2018 & iNat 2019 & iNat 2021-mini \\
\toprule    
BYOL    & 69.8 {\scriptsize(68.6{±}0.9)} & 77.4 {\scriptsize(76.7{±}0.8)} & 70.5 {\scriptsize({69.1{±}1.1})}\\
MoCo v3 & 70.1 {\scriptsize(69.4{±}0.5)} & 77.6 {\scriptsize(77.2{±}0.4)} & 70.9 {\scriptsize({70.5{±}0.5})}\\
DINO$^\dagger$    & 72.1 {\scriptsize(71.9{±}0.2)} & 79.4 {\scriptsize(79.0{±}0.4)} & 73.0 {\scriptsize({72.8{±}0.1})}\\
iBOT$^\dagger$    & 73.8 {\scriptsize({73.5{±}0.2})} & 79.9 {\scriptsize(79.5{±}0.4)} & 74.5 {\scriptsize({74.4{±}0.1})}\\
data2vec &  75.2 {\scriptsize({74.5{±}0.7})} & 80.6 {\scriptsize(80.0{±}0.5)}  & 76.2 {\scriptsize(75.5{±}0.9)} \\
MAE  & 74.6 {\scriptsize(74.5{±}0.1)} & 80.2 {\scriptsize(80.0{±}0.1)} & 75.7 {\scriptsize({75.5{±}0.2})}\\
\midrule
\ours  & \bf{75.8 {\scriptsize(75.3{±}0.3)}} & \bf{81.0 {\scriptsize(80.5{±}0.4)}} & \bf{76.7 {\scriptsize({76.3{±}0.3})}}\\
\bottomrule
\end{tabular}
}
\vspace{-1.5em}
\end{table}

\begin{table*}[t]
\small
\caption{
\textbf{Transfer learning results on FGVC.} We present the end-to-end fine-tuning accuracies on multiple datasets, reporting the best results along with the mean ± std of the accuracies from grid searches. Our method mostly outperforms the competitors at the best accuracies, further showcasing the robustness among different training hyper-parameters. $^\dagger$ denotes the models pre-trained using multi-crop augmentation.
}
\label{tab:transfer_cls}
\tabcolsep=0.1em
\centering
\vspace{-1em}
\resizebox{1.01\linewidth}{!}{
\begin{tabular}{l c c c c c c c |c}
\toprule
Method & Aircraft & Birds  & CUB-200 & CIFAR-10 & CIFAR-100 & Dogs & Flowers & Avg\\
\toprule    
DINO$^\dagger$ & 87.0 {\scriptsize(86.0{±}0.6)} & 83.9 {\scriptsize(83.4{±}0.5)}  & 85.1 {\scriptsize(84.9{±}0.3)} & 99.0 {\scriptsize(98.9{±}0.1)} & 91.3 {\scriptsize(90.7{±}0.5)}   & 84.8 {\scriptsize(84.6{±}0.3)} & 98.8 {\scriptsize(98.7{±}0.1)} & 90.0\\
iBOT$^\dagger$ & 87.3 {\scriptsize(86.7{±}0.6)} & 85.5 {\scriptsize(85.1{±}0.5)} & 85.9 {\scriptsize(85.5{±}0.3)}& \textbf{99.2} {\scriptsize(98.8{±}0.6)} &  \textbf{92.0 {\scriptsize(91.1{±}0.9)}} & 86.0 {\scriptsize(85.7{±}0.3)}  & \textbf{99.0 {\scriptsize(99.0{±}0.1)}} & \underline{90.7} \\
MAE & 88.1 {\scriptsize(87.3{±}0.9)} & 84.2 {\scriptsize(84.0{±}0.3)} & 84.6 {\scriptsize(84.3{±}0.2)} & 98.8 {\scriptsize(98.7{±}0.1)} &  90.0 {\scriptsize(89.7{±}0.3)} & 86.8 {\scriptsize({86.4±0.3})} & 98.1 {\scriptsize(97.8{±}0.3)} & 90.1\\
data2vec &  87.3 {\scriptsize(86.6{±}0.7)} & 84.1 {\scriptsize(83.5{±}0.5)} & 84.4 {\scriptsize(83.9{±}0.4)} & 98.8 {\scriptsize(98.7{±}0.1)}  & 91.2 {\scriptsize(91.0{±}0.2)} &  85.7 {\scriptsize(85.3{±}0.3)}  & 96.7 {\scriptsize(94.4{±}3.3)} & 89.7\\
\midrule
 \ours  & \bf{89.2 {\scriptsize(88.3{±}0.9)}} & \bf{86.0 {\scriptsize(85.3{±}0.6)}} & \bf{86.5 {\scriptsize({85.7±0.6})}} & 99.1 \textbf{\scriptsize(98.9{±}0.1)} & 91.0 {\scriptsize(90.7{±}0.4}) & \bf{87.4 {\scriptsize(86.7{±0.5})}} & 98.4 {\scriptsize(98.2{±}0.2)} & \textbf{91.1}\\
\bottomrule
\end{tabular}
}
\vspace{-1em}
\end{table*}

\subsection{Transfer Learning}
\subsubsection{iNaturalist datasets.}

To further compare the transferability of learned representations, we measure image classification accuracies by fine-tuning on iNaturalist 2018, iNaturalist 2019, and mini iNaturalist 2021~\cite{inaturalist}, which are highly imbalanced with different number of images per class. We compare \ours with MoCo v3~\cite{chen2021mocov3}, BYOL~\cite{grill2020byol}, DINO~\cite{caron2021emerging}, iBOT~\cite{zhou2021ibot}, and MAE~\cite{mae}. All the models are ImageNet-1k pre-trained ViT-B/16 with a resolution of 224$\times$224. We report the maximum accuracy with the mean and standard deviation of the accuracies obtained by grid searches of learning rates and weight decay, following the protocol~\cite{kornblith2019do_imagenet}. Table~\ref{tab:transfer_cls_more} shows \ours outperforms the competitors across all datasets, which reveals superior transferability and tuning robustness.

\vspace{-1em}
\subsubsection{Fine-Grained Visual Classification (FGVC) datasets.}
We further validate fine-tuning classification accuracies on CIFAR-10~\cite{cifar}, CIFAR-100~\cite{cifar}, CUB-200~\cite{cub-200}, Aircraft~\cite{fine-grained}, Birds~\cite{nabirds}, Flowers~\cite{oxfordflower}, and Dogs~\cite{stanforddogs} following the same evaluation protocol as above. Table~\ref{tab:transfer_cls} showcases \ours achieves the best number on average and outstanding numbers overall, which shows improved transferability and tuning robustness across datasets again.

\section{Analysis and Discussion}
\label{sec:analysis}

\subsection{Ablation Study}\label{sec5:analysis}
We conduct ablation studies of \ours pre-training under various available configurations. We use ViT-B/16 and train it for 400 epochs on ImageNet-1K as the fixed pre-training setup. Each model is then individually pre-trained. We report the top-1 fine-tuning and linear probing accuracies for each study.

\noindent\textbf{Steered tokens.}
We showed that steering unmasked tokens via comprehensive supervision leads to an improved encoder. Additionally, we investigate whether masked tokens also benefit from steering. Table~\ref{tab:guided_token} illustrates that training solely with unmasked tokens shows superiority, aligning with our primitive conjecture.

\noindent\textbf{Contextualization for token steering.}
We mainly used the visual tokens for contextualizing methods, but we studied whether other tokens, such as cls-token, can be an alternative in Table~\ref{tab:ablation_token}. We observe using pooled visual tokens is preferred for \ours. Considering latent features undergo masked auto-encoding, these results imply that explicitly using general context is more effective than using implicit information via cls-token.

\noindent\textbf{Type of supervision.}
We study the effectiveness of various supervisions for unmasked tokens in Table~\ref{tab:ablation_type}. We mainly compare token-wise versus broadly aggregated supervision. While all the types yield performance gains, sole contextualization works the best. It outperforms the combination of token-wise, implying that the additional token-wise supervision may conflict with the aggregated one, which is presumably due to the alignment between the set of tokens.

\noindent\textbf{Loss function.} \label{sec4:varying_ftn}
We explore various losses for the broader contextualization loss in Table~\ref{tab:ablation_ftn}. While all objectives yield considerable performance as expected above, the cosine distance of latent representations of broader and partial information works best when pre-training by \ours.

\noindent\textbf{Masking ratio for target.} \label{sec4:varying_target_mask} We study whether the target encoder needs masked images. Table~\ref{tab:ablation_target_mask} shows that the model without target masking outperforms all the counterparts. Moreover, the fine-tuning accuracy with masking even underperforms the baseline, implying that transferring coarse information harms the capability of learning representation.

\noindent\textbf{Image crop type.} \label{sec4:varying_crop}
This study highlights how performance is affected by the disparity between the two views. There would be many comparing options; we choose Random resized crop (RRC)~\cite{inceptionv1} and simple resized crop (SRC)~\cite{touvron2022deit} for comparison. Table~\ref{tab:varying_crop} shows the model pre-trained with SRC exceeds the fine-tuning accuracy of the case of RRC. Since RRC is more compatible with MAE than SRC, performance improvements are not observed in MAE. Our method benefits from SRC, which indicates that the information that will be encoded needs to align closely with the view of the other side, thereby facilitating training.

\begin{table*}[t!]
\centering
\small
\caption{\textbf{Ablation studies}. We report fine-tuning (ft) and linear probing (lin) accuracies for each configuration which are pre-trained with ViT-B/16.
All models are pre-trained for 400 epochs on ImageNet-1k. We mark our default settings in \colorbox{baselinecolor}{gray}.}
\label{tab:ablation}
\subfloat[
\textbf{Steered tokens}. Guiding unmasked tokens only performs better. 
\label{tab:guided_token}
]{
\tabcolsep=0.15cm
\centering
\begin{minipage}{0.42\linewidth}
\resizebox{1\textwidth}{!}{
\begin{tabular}{lcc}
case & ft & lin  \\
\toprule
none & 82.8 & 61.5 \\
unmasked/masked tokens & 83.0 & 67.2 \\
unmasked tokens only & \baseline{\textbf{83.5}} & \baseline{\textbf{67.9}} \\
\end{tabular}
}
\end{minipage}
}
\subfloat[
\textbf{Context tokens.}
Using visual tokens works better.
\label{tab:ablation_token}
]{

\begin{minipage}{0.42\linewidth}
\centering
\tabcolsep=0.15cm
\resizebox{.89\textwidth}{!}{
\begin{tabular}{lcc}
case & ft & lin  \\
\toprule
- & 82.8 & 61.5 \\
cls-token & 83.2 &  \textbf{71.0} \\
pooled visual tokens  & \baseline{\textbf{83.5}} & \baseline{67.9} \\
\end{tabular}
}
\end{minipage}
}
\\
\subfloat[
\textbf{Supervision types}. 
Contextualization beats others.
\label{tab:ablation_type}
]{
\begin{minipage}{0.42\linewidth}
\centering
\tabcolsep=0.15cm
\resizebox{.83\textwidth}{!}{
\begin{tabular}{lcc}
case & ft  &  lin  \\
\toprule
none & 82.8 & 61.5 \\
token-wise & 83.2  & 64.1   \\
contextualization & \baseline{\textbf{83.5}}  &   \baseline{\textbf{67.9}} \\
both supervisions & 83.1  & 66.6   \\
\end{tabular}
}
\end{minipage}
} 
\hspace{1em}
\subfloat[
\textbf{Loss function}. %
``Cos'' works the best.
\label{tab:ablation_ftn}
]{
\centering
\begin{minipage}{0.42\linewidth}
\centering
\vspace{-0.5em} 
\tabcolsep=0.15cm
\resizebox{.77\textwidth}{!}{
\begin{tabular}{lcc}
case & ft & lin \\
\toprule
none & 82.8  & 61.5 \\
infoNCE & 83.0  & 66.7   \\
smoothed-$\ell_1$ & 83.2  & 60.2  \\
cosine distance & \baseline{\textbf{83.5}}  &  \baseline{\textbf{67.9}} \\
\end{tabular}
}
\end{minipage}
}
\\
\vspace{-1.5em}
\subfloat[
\textbf{Target's masking ratio}. Target encoder needs noncorrupted views. %
\label{tab:ablation_target_mask}
]{
\begin{minipage}{0.42\linewidth}
\centering
\tabcolsep=0.15cm
\resizebox{.97\textwidth}{!}{
\begin{tabular}{lccc}
case & ratio & ft & lin \\
\toprule
none (w/o target) &  -  &  82.8 & 61.5 \\
w/o target masking & 0  &  \baseline{\textbf{83.5}}    &  \baseline{\textbf{67.9}} \\
w/ target masking & 0.5 &  82.6 & 63.2 \\
w/ target masking & 0.75 & 82.5 & 63.8 \\
\end{tabular}
}
\end{minipage}
}
\hspace{1em} 
\subfloat[
\textbf{View discrepancy}. Ours benefits reduced differences among views. \label{tab:varying_crop} 
]{
\begin{minipage}{0.42\linewidth}
\tabcolsep=0.15cm
\centering
\vspace{.5em} 
\resizebox{1\textwidth}{!}{
\begin{tabular}{lccccc}
\multirow{2}{*}{Method}  & \multirow{2}{*}{Epochs} & \multicolumn{2}{c}{Views} & \multirow{2}{*}{ft}  & \multirow{2}{*}{lin}\\
  & & RRC & SRC & &   \\
\toprule
MAE & 400 & \checkmark & & 82.8  & 61.5 \\
MAE & 400 &  & \checkmark & 82.5 & 64.2 \\
\midrule
\ours & 400 & \checkmark  & & 83.4 & 67.1\\
\ours & 400 &   & \checkmark  & \baseline{\textbf{83.5}} & \baseline{\textbf{67.9}}\\
\bottomrule
\end{tabular}
}
\end{minipage}
}
\vspace{-2.5em}
\end{table*}

\subsection{Further Analyses}

\subsubsection{Grad-CAM visualizations.}
\setlength{\columnsep}{4pt}%
\begin{wrapfigure}{r}{0.44\columnwidth}
\vspace{-4.8em}
\centering
\begin{minipage}{0.12\columnwidth}
\centering
\includegraphics[width=\columnwidth]{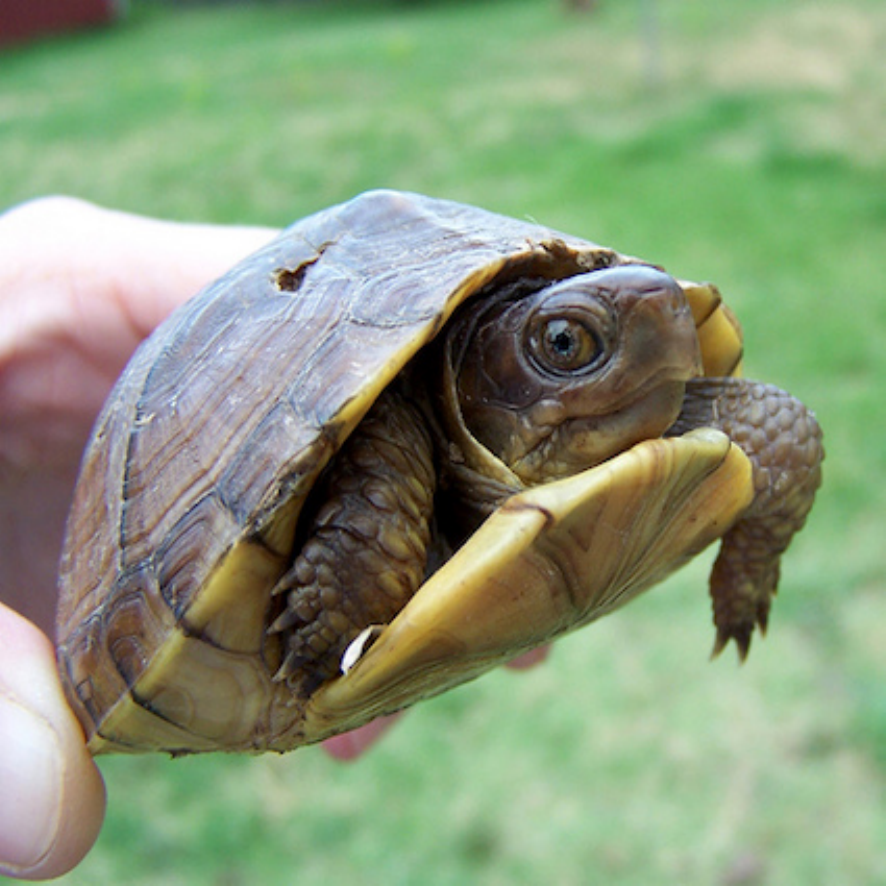}
\end{minipage}
\begin{minipage}{0.12\columnwidth}
\centering
\includegraphics[width=\columnwidth]{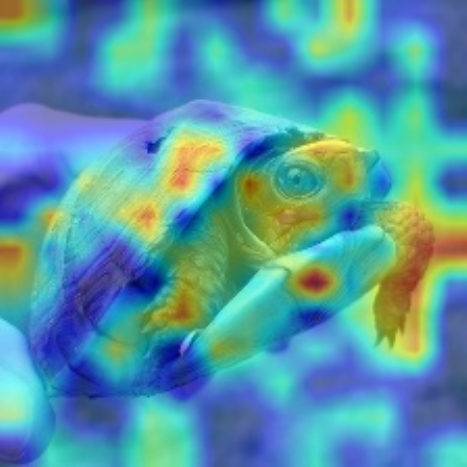}
\end{minipage}
\begin{minipage}{0.12\columnwidth}
\centering
\includegraphics[width=\columnwidth]{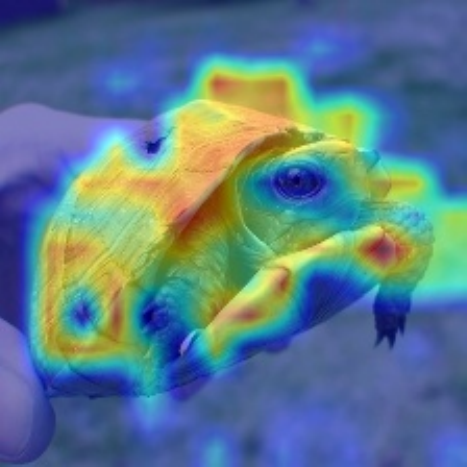}    
\end{minipage}\\
\hspace{.02em}
\begin{minipage}{0.12\columnwidth}
\centering
\includegraphics[width=\columnwidth]{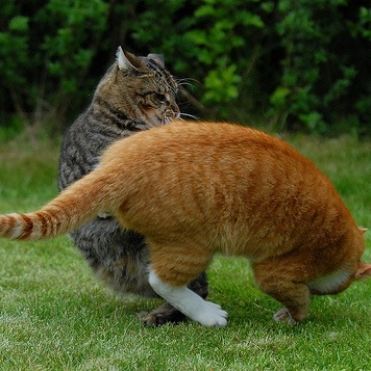}
\subcaption{{\small Input}}
\end{minipage}
\begin{minipage}{0.12\columnwidth}
\centering
\includegraphics[width=\columnwidth]{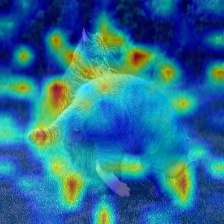}
\subcaption{{\small MAE}}
\end{minipage}
\begin{minipage}{0.12\columnwidth}
\centering
\includegraphics[width=\columnwidth]{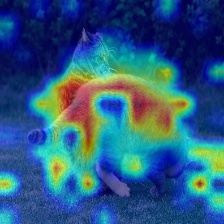}    
\subcaption{{\small Ours}}
\end{minipage}
\vspace{-1em}
\caption{\textbf{GradCAM visualization.}}%
\label{fig:GRADCAM}
\vspace{-4em}
\end{wrapfigure}

We analyze the Grad-CAM results for \ours vs. MAE to assess the impacts of our method in terms of context range. Fig.~\ref{fig:GRADCAM} shows \ours better grasps broader foreground object contexts, while MAE has difficulty in fully emphasizing these regions.

\begin{figure}[t]
    \centering
\begin{minipage}[t]{0.48\linewidth}
    \centering
    \includegraphics[width=0.9\linewidth]{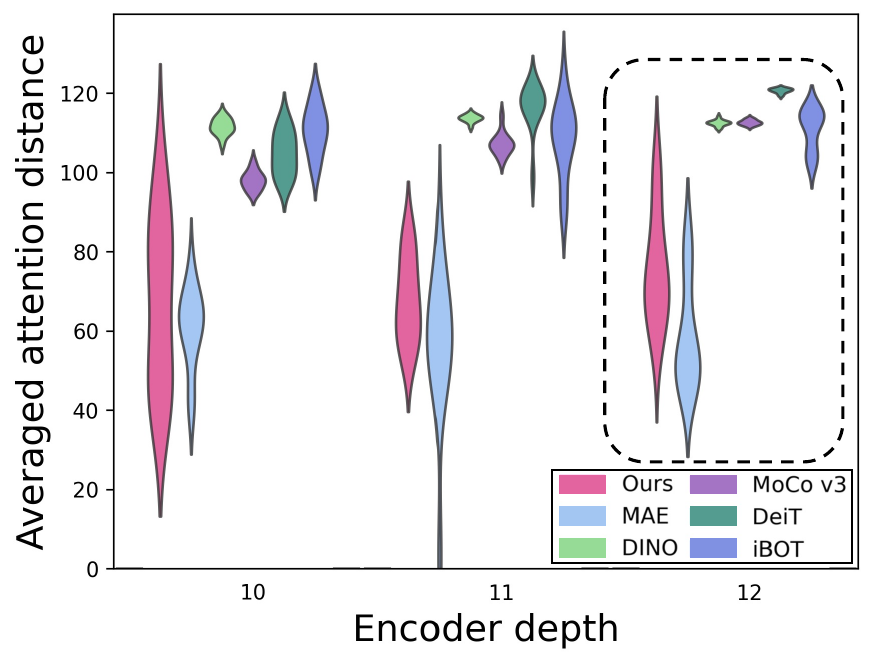}
    \subcaption{Averaged attention distance (AAD)}\label{fig:aad}
\end{minipage}
\hfill
\begin{minipage}[t]{0.48\linewidth}
    \centering
    \includegraphics[width=0.9\linewidth]{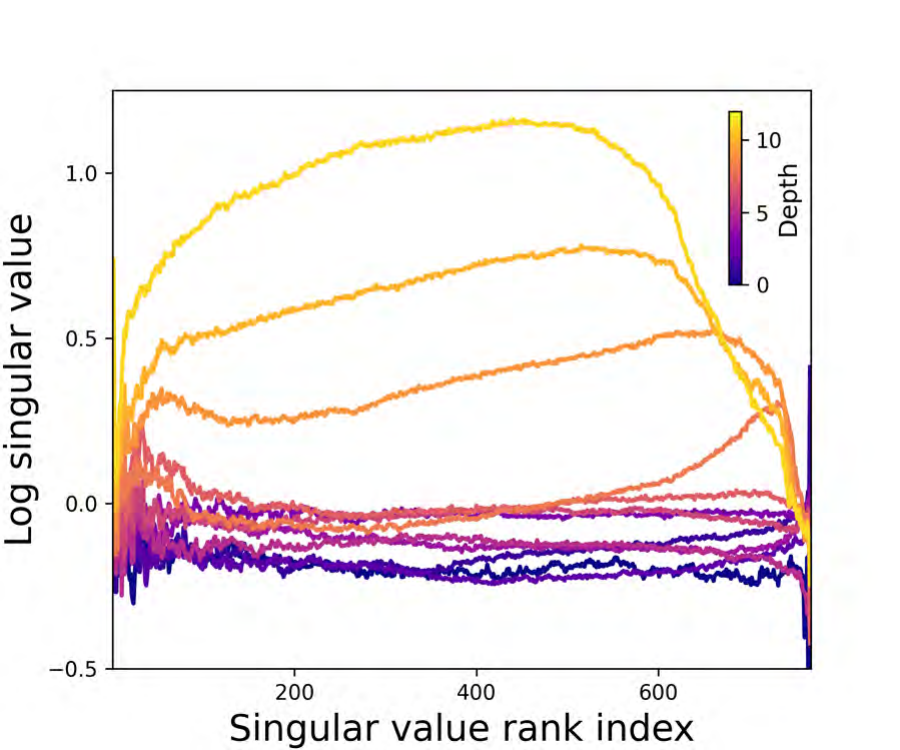}
    \subcaption{Singular value (SV) spectrums}
\end{minipage}
\vspace{-.5em}
    \caption{\textbf{Visualization Results}. 
    (a) We plot AAD values of our \ours, MAE \cite{mae}, DINO~\cite{caron2021emerging}, MoCo v3 \cite{chen2021mocov3}, DeiT \cite{touvron2021deit}, and iBoT~\cite{zhou2021ibot} for the last three layers, which handle high-level semantics. \ours ranges more diverse and broader dependencies overall.
    (b) We plot the difference of singular values between the baseline and \ours at each layer, showing large gaps (${\geq}0$), particularly for the later layers. This suggests our method learns more discriminative representations.
    }\label{fig:svs_gap_ours_mae}
    \vspace{-1em}
\end{figure}

\vspace{-1em}
\subsubsection{Analysis of attention distance.}
We measure the average attention distance (AAD)~\cite{dosovitskiy2020vit} to explore the dependency range in MAE quantitatively. Formally, AAD is defined by a given self-attention map $S$ as follows:
\begin{equation}
\label{eq:aad}
    d_{\text{AAD}} = \sum_{j\in\mathcal{M}} \sum_{i\in\mathcal{M}} S(i,j) \cdot ||p_i - p_j||_2,
\end{equation}
where $\mathcal{M}$ denotes a set of patch indices and $S(i,j)$ denotes an attention value between $i$-th and $j$-th patches. $p_i$ is a normalized 2D coordinate of the center of the $i$-th patch. We compute AAD using the entire images from the ImageNet-1K validation set. ViT-B/16 pre-trained by MAE \cite{mae}, DINO~\cite{caron2021emerging}, MoCo v3 \cite{chen2021mocov3}, DeiT \cite{touvron2021deit}, and iBoT~\cite{zhou2021ibot}on ImageNet-1K are included in our study. Considering the crucial role of the final layers in the semantic encoding process, our analysis focuses on the layer-wise AAD for the last three layers. Figure~\ref{fig:aad} shows that our \ours exhibits diverse ranges, spanning from shorter to longer-range dependencies. Thus, our model generally interacts wider than MAE and also surpasses others in the distance scopes. Notably, the phenomenon is more significant at the final layer, which determines the capacity of  contributes most to high-level semantics.

\vspace{-1em}
\subsubsection{Spectral analysis.}%
We provide additional analysis on the learned layer-wise representations \ours and MAE. Inspired by the previous study \cite{dimensional_collapse}, we measure the singular values (SVs) of the covariance of features, \ie, how the features are spread in the embedding space. More specifically, we compute a feature covariance matrix on ImageNet-1K validation set (\ie, the covariance matrix has a shape of 50k${\times}$50k), and compute the SVs of the covariance matrix. 
Fig.~\ref{fig:svs_gap_ours_mae} shows a spectrum of log of singular value gaps between MAE and \ours across the layers.
The singular values of \ours surpass the values of MAE across the rank indices in the last layers, while both methods have similar singular values on earlier layers.
The results reveal that \ours have larger singular values at the output-side layers, indicating a higher rank of the feature space~\cite{yang2017breaking,han2021rethinking}. In other words, \ours utilizes the output feature space better than MAE, owing to the longer-range understanding prompted by broader contextualization.\\

\begin{table*}[t]
    \centering
    \caption{\textbf{Robustness evaluation.} We evaluate the robustness of the representative ImageNet-1K pre-training methods: DINO, iBOT, MAE, and data2vec with our \ours on in-distribution generalization ({IN-V2/Real}) and out-of-distribution ({IN-A/IN-O/Sketch/R/Cocc/Obj}) benchmarks. We also evaluate the capability to detect spurious correlations with background on SI-Score metrics~\cite{si_score}. We highlight the best numbers (in boldface) and the second-best numbers (in underlined). $^\dagger$ denotes the models pre-trained using multi-crop augmentation. %
    Ours surpasses others than expected by the ImageNet-1K numbers, particularly more on localization-related metrics.%
    }
    \label{tab:robust_brief} 
    \vspace{-1em}
    \small
    \tabcolsep=0.2em    
    \resizebox{1.0\linewidth}{!}{
    \begin{tabular}{@{}lccccccccccccc@{}}
        \toprule
         & \scriptsize IN-1k↑ & \scriptsize IN-V2↑ & \scriptsize IN-Real↑ & \scriptsize IN-A↑ & \scriptsize IN-O↑ &\scriptsize  Sketch↑ & \scriptsize IN-R↑ & \scriptsize Cocc↑ & \scriptsize ObjNet↑ & \scriptsize SI-size↑ & \scriptsize SI-loc↑ & \scriptsize SI-rot↑ \\
        \toprule
        DINO$^\dagger$ & 83.1 & 72.8 & 87.6 & 36.3 & 60.7 & 35.7 & 48.2 & 77.8 & 36.4 & 57.8 & 37.0 & 43.8 \\
        IBOT$^\dagger$ & 83.5 & 73.5 & 87.9 & 39.4 & 62.0 & 37.8 & 50.2 & 78.6 & 37.1 & \underline{58.2} & 37.6 & \underline{43.9}  \\
        MAE  & 83.7 & 72.9 & 88.2 & 36.7 & \textbf{65.4} & 35.9 & 48.9 & 78.4 & 37.6 & 58.0 & \underline{38.7} & 42.7 \\
        data2vec & \underline{84.1} & \underline{74.2} & \underline{88.5} & \underline{41.6} & 62.2 & \textbf{38.7} & \textbf{53.0} & \underline{79.1} & \textbf{40.3} & 57.9 & 38.6 & 43.8 \\
        \midrule
        \ours & \textbf{84.2} & \textbf{74.2} & \textbf{88.6} & \textbf{42.5} & \underline{64.1} & \underline{38.2} & \underline{52.1} & \textbf{79.2} & \underline{38.9} & \textbf{59.8} & \textbf{40.7} & \textbf{44.9} \\
        \bottomrule
    \end{tabular}
    }
    \vspace{-1.5em}
\end{table*}

\vspace{-1em}
\subsection{Robustness Evaluation}
We evaluate the robustness of various methods, including DINO~\cite{caron2021emerging}, iBOT~\cite{zhou2021ibot}, MAE~\cite{mae}, and data2vec~\cite{data2vec} with \ours on various robustness benchmarks. We can interpret how our method impacts model robustness. 
We employ two in-distribution benchmarks including ImageNet-V2~\cite{imagenet_v2} and -Real~\cite{are_we_done}) and four out-of-distribution benchmarks ImageNet-A~\cite{hendrycks2021imagenet_a}, -O~\cite{hendrycks2021imagenet_a}, -R~\cite{hendrycks2021imagenet_r}, -Sketch~\cite{imagenet_sketch}, and ObjectNet~\cite{objectnet}. We further use SI-Score~\cite{si_score} %
to test spurious correlations with the background. Lastly, we evaluate the center occlusion benchmark that zeroes the center patch in the ImageNet-1K evaluation images. As shown in Table~\ref{tab:robust_brief}, \ours achieves outstanding performance on all the benchmarks.
\section{Conclusion}
We have introduced a novel framework to address the limited broader understanding of images inherent in MIM. We argued that MIM, such as MAE, learns a narrower range of dependency due to lacking a comprehensive understanding of entire pixels. 
By visualizing attention maps, we have shown that MAE exhibits incomplete coverage of foreground or background regions. We conjectured this is caused by the potential absence of general context in learned unmasked tokens when interacting with mask tokens in self-attentions. Based on the observation, we have proposed \ours pre-training method, minimizing the discrepancy between the context features and sparse visual tokens through our broader contextualization loss. Our simple remedy enhanced MAE by a large margin on ImageNet-1K and ADE20K, and enables \ours to be very competitive to state-of-the-art methods. \ours further offers significant improvements in transfer learnings, including the iNatrualist and FGVC datasets.
Finally, measuring averaged attention distance and spectral analysis demonstrated that \ours can be a simple yet effective supplement for masked image modeling.

\noindent\textbf{Limitations.} We verified our method's applicability only up to ViT-L. Resource constraints prevented us from performing more experiments on larger models. We will further confirm the scalability with ViT-G or ViT-H.

\bibliographystyle{splncs04}
\bibliography{main}

\clearpage
{\noindent \large \textbf{Appendix}}
\appendix
\numberwithin{equation}{section}
\numberwithin{figure}{section}
\numberwithin{table}{section}

\renewcommand\thefigure{\Alph{figure}}    
\setcounter{figure}{0}  
\renewcommand\thetable{\Alph{table}}    
\setcounter{table}{0} 

\vspace{2em}

\noindent This appendix includes additional experimental analyses of our proposed method, comparing it with state-of-the-art self-supervised learning (SSL) methods and experimental results with detailed setups.
We first provide the attention map visualizations in \S\ref{sec:attention_map}; we then provide 1) another applicability of our proposed method to SimMIM, 2) ablation studies, and 3) our implementation details, including hyper-parameters in \S\ref{sec:further_exp}.

\section{On Distinctiveness of Attention Map}
\label{sec:attention_map}
\begin{figure*}[p]
     \centering
     \begin{subfigure}{0.048\linewidth}
         \centering
         \includegraphics[width=\textwidth]{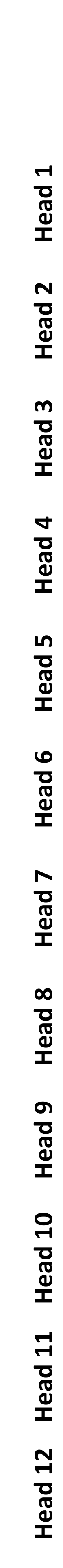}
         \label{fig:index}
     \end{subfigure}
     \begin{subfigure}[b]{0.12\linewidth}
         \centering
         \includegraphics[width=0.75\textwidth]{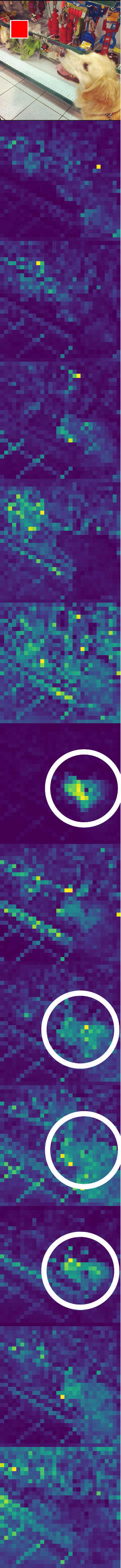}
         \caption{ DINO
         }
     \label{fig:attn_maps_all_heads_appendix_moco}
     \end{subfigure}
     \begin{subfigure}[b]{0.12\linewidth}
         \centering
         \includegraphics[width=0.75\textwidth]{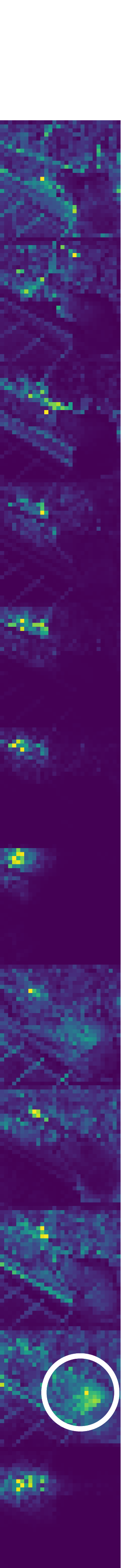}
         \centering
         \caption{iBOT 
         }
    \label{fig:attn_maps_all_heads_appendix_ibot}
     \end{subfigure}
     \begin{subfigure}[b]{0.15\linewidth}
         \centering
         \includegraphics[width=0.6\textwidth]{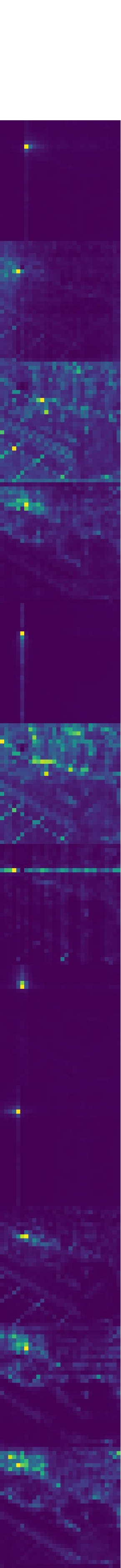}
         \caption{data2vec
         }
         \label{fig:attn_maps_all_heads_appendix_data2vec}
     \end{subfigure}
     \begin{subfigure}[b]{0.12\linewidth}
         \centering
         \includegraphics[width=0.75\textwidth]{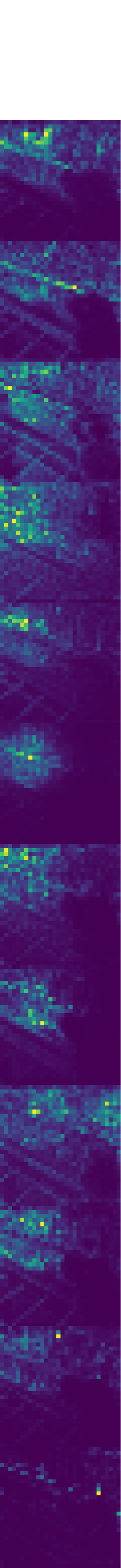}
         \caption{MAE
         }
         \label{fig:attn_maps_all_heads_appendix_mae}
     \end{subfigure}
     \begin{subfigure}[b]{0.12\linewidth}
         \centering
         \includegraphics[width=0.75
         \textwidth]{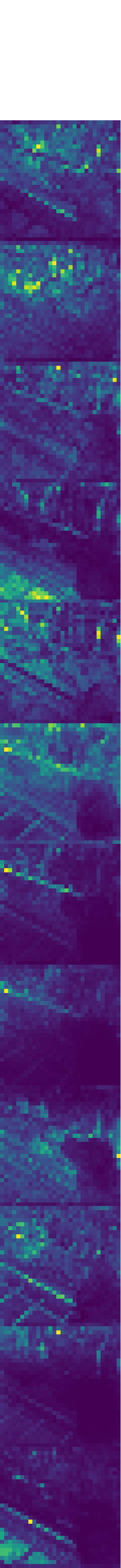}
         \caption{Ours
         }
         \label{fig:attn_maps_all_heads_appendix_ours}
     \end{subfigure}
    \caption{{\bf Attention visualization for all multi-heads} of the last self-attention block. Given a sample and a query (left top on Fig A.3(a)), We visualize the attention maps of the models (with ImageNet-1K accuracies) pre-trained by DINO~\cite{caron2021emerging}, iBOT~\cite{zhou2021ibot}, data2vec~\cite{data2vec}, MAE~\cite{mae}, and \ours. Each row presents the corresponding attention map of each head. White circles in the attention maps emphasize the highlighted foreground regions despite the background query. We use the ViT-B/16 architecture and a resolution of 224$\times$224. We borrowed a sample image from \texttt{n2099601} ImageNet-1K class. }
    \label{fig:attn_maps_all_heads_appendix}
\end{figure*}

\begin{figure*}[p]
     \centering
     \begin{subfigure}{0.048\linewidth}
         \centering
         \includegraphics[width=\textwidth]{materials/attn_all_heads_app/index_all_heads2.pdf}
         \label{fig:index2}
     \end{subfigure}
     \begin{subfigure}[b]{0.12\linewidth}
         \centering
         \includegraphics[width=0.75\textwidth]{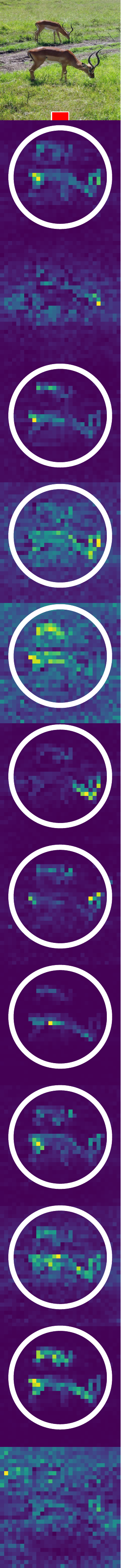}
         \caption{DINO
         }
         \label{fig:attn_maps_all_heads2_appendix_moco}
     \end{subfigure}
     \begin{subfigure}[b]{0.12\linewidth}
         \centering
         \includegraphics[width=0.75\textwidth]{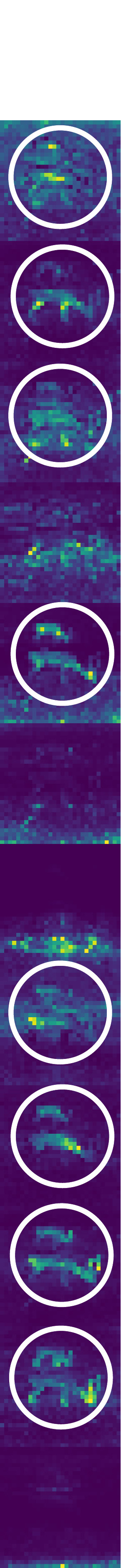}
         \caption{iBOT
         }
         \label{fig:attn_maps_all_heads2_appendix_ibot}
     \end{subfigure}
     \begin{subfigure}[b]{0.15\linewidth}
         \centering
         \includegraphics[width=0.6\textwidth]{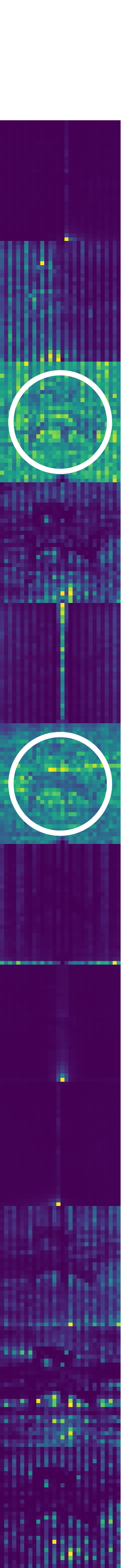}
         \caption{data2vec
         }
         \label{fig:attn_maps_all_heads2_appendix_data2vec}
     \end{subfigure}
     \begin{subfigure}[b]{0.12\linewidth}
         \centering
         \includegraphics[width=0.75\textwidth]{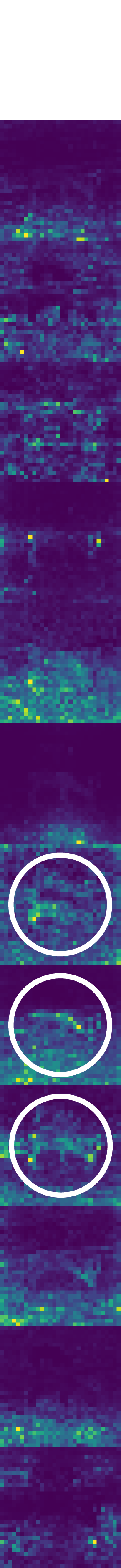}
         \caption{MAE
         }
         \label{fig:attn_maps_all_heads2_appendix_mae}
     \end{subfigure}
     \begin{subfigure}[b]{0.12\linewidth}
         \centering
         \includegraphics[width=0.75\textwidth]{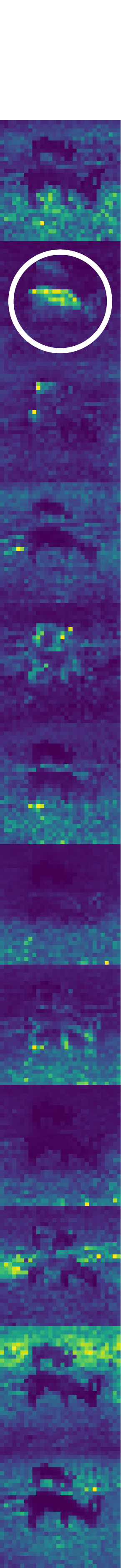}
         \caption{Ours 
         }
         \label{fig:attn_maps_all_heads2_appendix_ours}
     \end{subfigure}
        \caption{{\bf Attention visualization for all multi-heads} of the last self-attention block. Given a sample and a query (left top on Fig A.3(a)), We visualize the attention maps of the models (with ImageNet-1K accuracies) pre-trained by DINO~\cite{caron2021emerging}, iBOT~\cite{zhou2021ibot}, data2vec~\cite{data2vec}, MAE~\cite{mae}, and \ours. Each row presents the corresponding attention map of each head. White circles in the attention maps emphasize the highlighted foreground regions despite the background query. We use the ViT-B/16 architecture and a resolution of 224$\times$224. We borrowed a sample image from \texttt{n2422699} ImageNet-1K class. The grid pattern in (c) is presumably induced by the interpolation of the relative pose bias.}
        \label{fig:attn_maps_all_heads2_appendix}
\end{figure*}

In this section, we qualitatively show the improved discriminative power of our model compared with other SSL methods~\cite{caron2021emerging, zhou2021ibot, data2vec, mae} and \ours through attention map visualization by visualizing all the multi-heads of the last self-attention block using sample cases. We visualize the attention maps of the entire heads of the last self-attention according to the given query patches. We compare the diverse methods to investigate the distinctive trends. Fig.~\ref{fig:attn_maps_all_heads_appendix} and Fig.~\ref{fig:attn_maps_all_heads2_appendix} showcase when the input queries are from the background of the images,
As shown in Fig.~\ref{fig:attn_maps_all_heads_appendix}, models pre-trained with %
DINO~\cite{caron2021emerging} highlight foreground regions despite the background query, which reveals %
DINO broadly aggregates representations across the image, losing discriminative power. Moreover, iBOT also suffers from the correlation between the representations of foreground and background patches, as observed in Fig.~\ref{fig:attn_maps_all_heads_appendix_ibot} and Fig.~\ref{fig:attn_maps_all_heads2_appendix_ibot}.
data2vec shows precise local discriminatibility in Fig.~\ref{fig:attn_maps_all_heads_appendix_data2vec}, but indiscriminatively highlights attention in Fig.~\ref{fig:attn_maps_all_heads2_appendix_data2vec}.
While MAE does not confuse foreground and background representations in Fig.~\ref{fig:attn_maps_all_heads_appendix_mae}, MAE suffers another confusion in Fig.~\ref{fig:attn_maps_all_heads2_appendix_mae}, which may stem from lack of broader contexts. Besides, \ours shows enhanced discriminability between foreground and background patches in both cases.

\section{Experiments (cont'd)}
\label{sec:further_exp}
This section presents continued experiments that further investigate the superiority and applicability of our method. We show another application of broader context supervision in masked image modeling beyond MAE. We finally share our experimental regimes for the ImageNet-1k fine-tuning and semantic segmentation experiments on ADE20K.

\begin{table}[t]
\small
\centering
\caption{
\textbf{Impact of broader contextualization in SimMIM}.
To verify the versatility of our method to other methods, we apply the proposed broader contextualized supervision to training SimMIM. All models are pre-trained and fine-tuned on ImageNet-1K. We employ ViT-B/16 trained with the image resolution of 224$\times$224 and the identical weighting parameter of 0.25 for our context supervision loss (\ie, $\mathcal{L}_{\text{BC}}$).
}
\label{tab:simmim_sg}
\begin{tabular}{@{}l c c }
\toprule
Method & Pre-training epochs & Accuracy ($\%$) \\
\toprule    
SimMIM & 100 & 81.6 \\
LUT (SimMIM) & 100 & \bf{81.8} \\
\bottomrule
\end{tabular}
\end{table}

\subsection{Further Applicability of Our Method}
We showcase another use case of our method with another baseline. We choose a representative masked image modeling SimMIM~\cite{xie2021simmim}. We aim to reveal that our solution is also compatible with other masked image modeling methods that do not drop mask tokens in the encoder, such as SimMIM~\cite{xie2021simmim}. 

We pre-train the models with SimMIM, which is the baseline, and SimMIM with our method on ImageNet-1K~\cite{imagenet} for 100 epochs and fine-tuned following the fine-turning recipe of SimMIM~\cite{xie2021simmim}. We primitively replace the masked image modeling part of our framework for MAE with SimMIM and employ the framework for training.
As shown in Table~\ref{tab:simmim_sg}, our method improves SimMIM by 0.2$\%$p despite short pre-training epochs, which shows the potential applicability of our method on MIMs.

\begin{table}[t]
\centering
\small
\tabcolsep=1.0em

\caption{\textbf{Loss balancing study}. We study the balance A weight between global guidance and MIM loss. All the studies report fine-tuning and linear probing accuracies for each configuration which are pre-trained with ViT-B/16.
All the backbones are pre-trained for 400 epochs. We mark the default settings for the study in \colorbox{baselinecolor}{gray}.}
\label{tab:loss_balancing}
\vspace{-1em}

\begin{center}
\begin{tabular}{l c c}
\toprule
Case & Fine-tuning ($\%$)   &  Linear probing ($\%$)\\
\toprule
0.1 & 83.2   & \textbf{70.7}  \\
\baseline{0.25} & \baseline{\textbf{83.5}}   &  \baseline{67.9} \\
0.5 & 83.4   & 70.1  \\
1.0 & 82.9   & 63.6 \\
\bottomrule
\end{tabular}
\end{center}
\vspace{-1em}
\end{table}

\begin{table}[t]
\centering
\small
\tabcolsep=1em
\caption{\textbf{Training with the Broader Contextualization loss (\ie, $\mathcal{L}_{\text{BC}}$) only}. 
All the models are pre-trained for 100 epochs on ImageNet-1K. Fine-tuned results on ImageNet-1K are reported.}
\label{tab:loss_bc_only}
\begin{center}
\vspace{-1em}
\begin{tabular}{l l}
\toprule
Method & Fine-tuning ($\%$) \\
\midrule
Baseline & 82.1  \\
$\mathcal{L}_{\text{BC}}$ only & 82.0 ($-$0.1\%p) \\
\midrule
LUT & 82.6 (+0.5\%p) \\
\bottomrule
\end{tabular}
\end{center}
\vspace{-1em}
\end{table}

\subsection{Balancing $\mathcal{L}_{\text{BC}}$}
To give a maximal impact through broader context supervision loss, we study an appropriate $\alpha$\ in Eq. (3), and Table~\ref{tab:loss_balancing} shows that a loss weight of 0.25 works best, and our method's effectiveness remains up to 0.5. 
Moreover, though the highly tilted loss weights brought relatively degraded performance, these models work better than a model pre-trained by MAE.

\subsection{How does training proceed when only using $\mathcal{L}_{\text{BC}}$?}
To further investigate its impact, we exclusively train with broader contextualization loss. We pre-train and fine-tune a ViT-B/16 on ImageNet-1K~\cite{imagenet}. As shown in Table~\ref{tab:loss_bc_only}, while the model pre-trained with $\mathcal{L}_{\text{BC}}$ results in on par accuracy to the baseline, which suggests a broad context decent supervision to the trainable encoder. However, it decreases the accuracy 0.6$\%$p from LUTs, demonstrating that the combination with the MIM loss learns more discriminative representations.

\begin{table}[h]
\small
\centering
\caption{\textbf{Hyper-parameter configurations for end-to-end fine-tuning on ImageNet-1K}. All the numbers are for fine-tuning with the ImageNet-1k pre-trained backbone to the ImageNet-1K classification.}
\label{tab:hyperparams_ft}
\begin{tabular}{y{135} | c}
\toprule
Config & Value \\
\midrule
Optimizer                   & AdamW \\
Base learning rate & 5e-4 (S), 2.5e-4 (B), 1e-3 (L)\\
Weight decay & 0.05 \\
Optimizer momentum & $\beta_1, \beta_2=0.9, 0.999$\\
Layer-wise learning rate decay & 0.75 (S), 0.65 (B, L) \\
Batch size                  &  1024   \\
Learning rate schedule & Cosine decay \\
Warmup epochs           & 5 \\
Training epochs             & 300 (S), 100 (B), 50 (L)\\
Resolution                  & $224\times 224$ \\
Augmentation & RandAug (9, 0.5)\\
Label smoothing & 0.1\\
Mixup & 0.8\\
Cutmix & 1.0 \\
Drop path & 0.1 \\
\bottomrule
\end{tabular}
\end{table}

\begin{table}[t]
\small
\centering
\caption{\textbf{Hyper-parameter configurations for the ADE20K finetuning}. All the numbers are for transfer learning with the ImageNet-1K pre-trained backbone to the ADE20K semantic segmentation.}
\label{tab:hyperparams_semseg}
\begin{tabular}{y{135} | c}
\toprule
Config & Value \\
\midrule
Optimizer                   & AdamW \\
Learning rate & 1e-4 \\
Weight decay & 0.05 \\
Optimizer momentum & $\beta_1, \beta_2=0.9, 0.999$\\
Layer-wise learning rate decay & 0.65 \\
Batch size                  &  16   \\
Learning rate schedule & Polynomial\\
Warmup iterations           & 1500 \\
Training epochs             & 160k\\
Resolution                  & $512\times 512$ \\
Drop path & 0.1\\
\bottomrule
\end{tabular}
\end{table}

\subsection{Additional Implementation Details}
\label{sec:imple}
\noindent\textbf{Fine-tuning setup for ImageNet-1K classification.}
We list the detailed hyper-parameters for fine-tuning on ImageNet-1K~\cite{imagenet} in Table~\ref{tab:hyperparams_ft}. Specifically, we use the AdamW optimizer and a weight decay 0.05 with a batch size of 1024. We used a layer-wise learning rate decay of 0.75 for ViT-S/16 and 0.65 for ViT-B/16 and ViT-L/16. We fine-tune ViT-S/16, ViT-B/16, and ViT-L/16 for 300, 100, and 50 epochs, respectively.\\

\noindent\textbf{Detailed setup for ADE20K semantic segmentation.}
We provide the detailed hyper-parameters for transfer learning 
to the semantic segmentation task on ADE20K~\cite{zhou2017ade20k} in Table~\ref{tab:hyperparams_semseg}. 
We fine-tune UperNet~\cite{upernet} initialized with our pre-trained model for 160k iterations with a batch size of 16. Note that we do not employ multi-scale training.

\end{document}